\documentclass[nopreprintline,12pt]{elsarticle}

\usepackage{amssymb}
\usepackage{amsmath}
\usepackage{amsthm}
\usepackage{color}
\usepackage{hyperref}
\usepackage{listings}
\usepackage{xcolor}

\journal{~}

\begin{document}

\begin{frontmatter}



\title{Language Models for Adult Service Website Text Analysis} 

\author[inst1]{Nickolas Freeman\corref{cor1}}
\ead{freem028@ua.edu}

\author[inst2]{Thanh Nguyen}
\ead{tpnguyen8@crimson.ua.edu}

\author[inst1]{Gregory Bott}
\ead{greg.bott@ua.edu}

\author[inst1]{Jason Parton}
\ead{jmparton@.ua.edu}

\author[inst2]{Collin Francel}
\ead{ctfrancel@crimson.ua.edu}

\cortext[cor1]{Corresponding author}

\affiliation[inst1]{Information Systems, Statistics, and Management Science, University of Alabama, Tuscaloosa, AL, USA}
\affiliation[inst2]{Data Science, University of Alabama, Tuscaloosa, AL, USA}


\begin{abstract}
Sex trafficking refers to the use of force, fraud, or coercion to compel an individual to perform in commercial sex acts against their will. Adult service websites (ASWs) have and continue to be linked to sex trafficking, offering a platform for traffickers to advertise their victims. Thus, organizations involved in the fight against sex trafficking often use ASW data when attempting to identify potential sex trafficking victims. A critical challenge in transforming ASW data into actionable insight is text analysis. Previous research using ASW data has shown that ASW ad text is important for linking ads. However, working with this text is challenging due to its extensive use of emojis, poor grammar, and deliberate obfuscation to evade law enforcement scrutiny. We conduct a comprehensive study of language modeling approaches for this application area, including simple information retrieval methods, pre-trained transformers, and custom transformer models. We demonstrate that characteristics of ASW text data allow efficient custom transformer models to be trained with relatively small GPU resources and used efficiently for inference on consumer hardware. Our custom models outperform fine-tuned variants of well-known encoder-only transformer models, including \texttt{BERT-base}, \texttt{RoBERTa}, and \texttt{ModernBERT}, on accuracy, recall, F1 score, and ROC AUC. We demonstrate the use of our best-performing custom configuration on three tasks related to ASW data analysis: ($i$) decomposing the giant component in a graph representation of ASW data, ($ii$) clustering ASW ad text, and ($iii$) using the learned token embeddings to understand the use of emojis in the illicit context we study. The models we develop represent a significant advancement in ASW text analysis, which can be leveraged in a variety of downstream applications and research.
\end{abstract}




\end{frontmatter}


\section{Introduction}\label{sec:introduction}
Sex trafficking involves the use of force, fraud, or coercion to compel an individual to perform commercial sex services. The internet facilitates sex trafficking in the US through Adult Services Websites (ASWs) that host commercial sexual advertisements (sex ads). To effectively combat this problem, law enforcement organizations (LEOs), non-profit organizations (NPOs), and researchers must transform sex ad data into actionable intelligence. Previous research using ASW data has shown that assessing the similarity of ASW ad text is important for linking ads. 

Given the recent emergence of powerful language models in Artificial Intelligence and Machine Learning (AI/ML), it seems plausible that many effective open-source model options for conducting such ASW ad text will already exist. However, our experience suggests that available models perform relatively poorly on this task. This poor performance stems from a significant mismatch between the training data used for these models and the text found in sex ads, which includes extensive emoji use, poor grammar, and deliberate obfuscation to evade law enforcement scrutiny.

We address the described performance gap by developing new models and techniques for understanding and comparing the similarity of texts collected from sex ads. Most research based on ASW data relies on smaller datasets that were sourced from Backpage, which ceased operations in 2018. We leverage a proprietary dataset that includes text from over 240 million sex ads collected from prominent ASWs between June 2020 and February 2025. To the best of our knowledge, this dataset represents the most extensive and pertinent sex ad dataset considered in the academic literature to date. We use this dataset to:
\begin{itemize}
    \item evaluate the effectiveness of both sparse encoding methods (TF-IDF variants) and dense encoding methods (transformer-based models) for representing ASW text,
    \item pre-train several new \texttt{BERT}-based foundation models specifically for ASW applications,
    \item empirically demonstrate that our custom pre-trained models outperform well-known models including \texttt{BERT-base}, \texttt{RoBERTa}, and \texttt{ModernBERT} when fine-tuned on ASW-specific datasets, and
    \item demonstrate how researchers and practitioners can use the developed models to ($i$) decompose the giant component that emerges in graph representations of ASW data, ($ii$) cluster sex ad text, and ($iii$) understand linguistic patterns in sex ads by analyzing the learned token embeddings.
\end{itemize}

\noindent The ASW-specific language models we propose provide researchers and practitioners with effective tools to transform sex ad data into intelligence that can be used to understand the online commercial sex ecosystem better and counter sexual exploitation facilitated by ASWs operating in this ecosystem.

\section{Related Work}\label{sec:related_work}
Our research addresses the challenge of generating actionable intelligence on potential sex trafficking victims using ASW data. To the best of our knowledge, we are the first to develop foundation models specifically for this context. The remainder of this section discusses related research on ($i$) analytical methods for transforming ASW data into actionable intelligence and ($ii$) the development of encoder-only foundation models. We focus on encoder-only models because our target application is author verification based on the full context of texts rather than text generation.

\subsection{Analytical Methods for Transforming ASW Data into Intelligence}\label{sec:related_work:transforming_asw_to_intelligence}
The literature on analytical methods for transforming ASW data into intelligence is diverse and continues to grow. This section focuses on recent works in this area that incorporate text analysis into their methods.

\subsubsection{Clustering and Pattern Detection}
Detecting clusters of related ASW data has been a significant research focus. \citet{li2018detection} describe an unsupervised template-matching algorithm for detecting patterns in ASW ad data that are indicative of organizations, which they use to refer to a single individual or a network of individuals who are trafficking a specific set of victims identified in ASW ads. Their template-matching approach uses GloVe embeddings to construct vector representations of text, which they use to cluster ads with HDBSCAN to obtain \textit{template signatures} (key phrases shared across clustered ads). They then use TF-IDF to construct graphs capturing the similarity between template signatures for potential merging. They provide examples showing their template-matching approach can identify textually similar ads effectively.

\citet{lee2021infoshield} present \texttt{INFOSHIELD}, a tool for identifying clusters of near-duplicate ads in ASW data. The tool uses TF-IDF to extract distinguishing keywords from posts and constructs a bipartite graph of document/phrase relationships. Connected components in the graph representation provide a \textit{coarse} data clustering, which is further refined in a \textit{fine} clustering phase. The authors validate their approach using Twitter bot data and the Trafficking10k dataset (see \citet{tong-etal-2017-combating} for more information on the Trafficking10k dataset). \citet{vajiac2023deltashield} extends this work to create a new method, \texttt{DELTASHIELD}, which is designed for ongoing analysis. 

More recently, \citet{nair2024t} propose \texttt{T-Net}, a 3-layer Graph Neural Network (GNN) for clustering ASW ads based on similar characteristics. This method uses \texttt{INFOSHIELD} to identify initial clusters of ads, which are represented as vertices in a graph. Each vertex in the graph has attributes related to the associated ads, such as the number of unique ads, phone number count, and location count. The authors apply different methods to label vertices exhibiting characteristics thought to be indicative of human trafficking.

\subsubsection{Named Entity Recognition}
Named Entity Recognition (NER) is another application area that has received substantial research interest. \citet{li2022extracting} introduce \texttt{NEAT} (Name Extraction Against Trafficking) for extracting person names in ASW data. To address the challenges posed by ambiguous names, the method combines rule-based and dictionary extractors with a fine-tuned \texttt{RoBERTa} model. The method is also able to adapt to adversarial changes in the text by expanding its dictionary. 

\citet{perez2023decoding} study the efficacy of language modeling and tokenization techniques for named entity recognition within the context of ASW data. Using a dataset of 1,810 annotated posts collected from a prominent ASW from January 2022 to April 2023, they find that the Longformer model with byte-level BPE tokenization performs best for entity extraction tasks.

\subsubsection{Trafficking Detection and Classification}
Several researchers have applied machine learning and artificial intelligence to detect trafficking indicators in ASW data directly. \citet{esfahani2019context} investigate the use of Latent Dirichlet Allocation (LDA), FastText word embeddings, and \texttt{BERT-base}, without any fine-tuning, to generate feature representations for classifying human trafficking in ASW ad text. Their training data consists of 10,000 ads collected from two ASWs in early 2017. They obtain labels for human trafficking via ``\textit{a database consisting of phone numbers associated with trafficking victims, constructed in conjunction with human trafficking domain experts.}'' Experiments suggest that combining all feature representations yields the best performance. 

\citet{zhu2019identification} train a normalized support vector machine (SVM) on TF-IDF representations of ASW ad text to classify ads into various risk categories. Their SVW models are trained using the Trafficking10k dataset. In a related but different application, \citet{li2023detecting} propose a natural language-based approach for classifying massage parlors as illicit based on Yelp reviews. They consider models that identify terms from a designed lexicon, models that generate embeddings, including \texttt{Doc2Vec} and \texttt{BERT}, as well as ensembles of lexicon- and embedding-based models. Although their experiments demonstrate trade-offs that make it hard to identify a \textit{best} model, they note that embedding-based models generally achieve higher recall values.

\subsubsection{Dataset and Resource Development}
Several researchers have developed and shared data and resources to facilitate research based on ASW data. \citet{saxena2023idtraffickers} present the \textit{IDTraffickers} dataset that contains 87,595 ASW ads collected between December 2015 and April 2016 from \textit{backpage.com}. The data includes 5,244 vendor labels obtained by grouping the ads based on phone numbers. The authors use the dataset to test various models for authorship identification and verification performance. They find the openly available \texttt{DeCLUTR-small} model (see \citet{giorgi2021declutrdeepcontrastivelearning}) performs best. However, as \citet{keskin2021cracking} demonstrates, relying solely on text and phone numbers for ad linkage is problematic. Their research shows that using perceptual hashing techniques to incorporate images into the linking process provides more comprehensive connections. This observation casts some doubt on the reliability of phone number-based vendor labels like those provided with \textit{IDTraffickers}.

More recently, \citet{Freeman-2025-multisite} describes a large, multi-site sample of ASW data that includes over 10 million advertisements collected from nine different ASWs between May 1, 2022, and August 1, 2022. To address privacy concerns, the authors obfuscate numerical data included in post texts, anonymized URLs and site names, and provide perceptual hash values instead of images.

\subsection{Encoder-only Foundation Model Development}\label{sec:related_work:encoder_only_foundation_models}
Language models have taken the world by storm since the release of ChatGPT. ChatGPT utilizes a Generative Pre-trained Transformer (GPT) foundation model to generate text based on input prompts. GPT models use only the decoder piece of the transformer architecture that is introduced in the seminal paper ``\textit{Attention is all you need}'' (see \citet{vaswani2017attention}). We focus on encoder-only architectures designed for bidirectional text understanding, because our goal is contextual text understanding rather than text generation. Specifically, our custom models build on the established \texttt{BERT-base} architecture, and we compare their performance to pre-trained variants of \texttt{BERT-base}, \texttt{RoBERTa}, and the more recent \texttt{ModernBERT}. This section provides an overview of these models, focusing on key implementation differences. We refer readers interested in additional details to the original papers referenced throughout.

\subsubsection{BERT}
\citet{devlin-etal-2019-bert} introduces the \texttt{BERT} family of models. The acronym \texttt{BERT} stands for \textit{Bidirectional Encoder Representations from Transformers}. \texttt{BERT} uses a WordPiece tokenizer to break words into smaller subword units. For example, \texttt{learning} becomes \texttt{learn} and \texttt{\#\#ing}, where the \texttt{\#\#} prefix indicates continuation from another subword. The original implementation uses a vocabulary of 30,522 tokens, comprising ($i$) 29,523 tokens corresponding to the most frequent subwords from BooksCorpus and English Wikipedia, ($ii$) five special tokens (\texttt{[PAD]}, \texttt{[UNK]}, \texttt{[CLS]}, \texttt{[SEP]}, and \texttt{[MASK]}), and ($iii$) 994 tokens \textit{unused} that allow customization. 

\texttt{BERT} is trained using two learning tasks: ($i$) \textit{masked language modeling} (MLM), where the model learns to predict randomly masked tokens in an input sequence based on surrounding tokens, and ($ii$) \textit{next sentence prediction} (NSP), where the model learns to predict sentences that follow others. During pre-training, 15\% of all WordPiece tokens are randomly masked in each sequence for the MLM task. This masking is performed once for all texts, and the same masking is used across all training epochs. 

The authors pre-train two \texttt{BERT} variants: ($i$) \texttt{BERT-base} and ($ii$) \texttt{BERT-large}. We base our custom models on \texttt{BERT-base}, which includes 12 transformer layers, 768-dimension hidden layers, and 12 self-attention heads that results in 110 million trainable parameters. The original pre-training uses the BooksCorpus (800M words) and English Wikipedia (2,500M words) datasets, a batch size of 256, and executed one million steps, corresponding to approximately 40 epochs over the pre-training corpus.

\subsubsection{RoBERTa}

\citet{roberta-2019} present \texttt{RoBERTA} (\textit{Robustly Optimized BERT pre-training Approach}), which improves upon \texttt{BERT} using enhanced training strategies and design implementations. Key modifications include ($i$) a new Byte-Pair Encoding (BPE) tokenizer with approximately 50,000 tokens, ($ii$) a tenfold increase in training data, ($iii$) removal of the NSP learning task, ($iv$) dynamic masking that re-samples throughout the training process, ($v$) larger batch sizes with more training steps, and ($vi$) optimized learning rate schedules and hyperparameters. \texttt{RoBERTa} outperforms \texttt{BERT} on several benchmarks, leading the authors to suggest that \texttt{BERT} was substantially undertrained and demonstrating the impact of training strategies on model performance.

\subsubsection{ModernBERT}
The rate of progress in encoder-only models has been much slower than that of decoder-only models in recent years due to the surge of interest in generative artificial intelligence. Recently, \citet{modernbert-2024} released \texttt{ModernBERT}, which uses a BPE tokenizer with a vocabulary size of 50,368. The model also includes several enhancements including: ($i$) embedding factorization that reduces the size of the embedding layer, ($ii$) alternating attention that changes between global and local attention in different layers, ($iii$) remove padding tokens and introduce a sequence concatenation approach to allow for more efficient in batch processing. The enhancements enable \texttt{ModernBERT} to handle sequences of 8,192 tokens efficiently on standard GPUs, while achieving state-of-the-art performance on various downstream tasks.

\subsection{Research Gap and Contributions}
As mentioned in Section \ref{sec:related_work:transforming_asw_to_intelligence}, research on analytical methods for combatting sex trafficking using ASW data is increasing, with many emerging studies recognizing the importance of ASW ad text. However, current approaches rely on sparse text representations (e.g., TF-IDF), simple dense representations (e.g., \texttt{Doc2Vec}), or pre-trained language models (e.g., \texttt{BERT} and \texttt{RoBERTa}). The absence of foundation models trained specifically on ASW data likely stems from two primary constraints: ($i$) limited publicly available data and ($ii$) significant computational requirements. Our research addresses these limitations through three key contributions:
\begin{enumerate}
    \item We use an unprecedented ASW dataset that includes over 19 million unique ad texts to develop and evaluate domain-specific foundation models optimized for authorship verification in illicit contexts.
    \item We demonstrate how to overcome computational resource challenges by exploiting distinct characteristics of ASW ad text, enabling efficient model training on standard hardware.
    \item We demonstrate the practical applications of our domain-specific models on three diverse tasks: ($i$) filtering the giant component that emerges in graph representations of ASW data, ($ii$) clustering ASW ads, and ($iii$) understanding the relationships between emojis commonly used in ASW ads.
\end{enumerate}

By developing the first foundation models tailored for ASW text analysis, this research lays the groundwork for more effective computational approaches for understanding the broad ASW ecosystem and combatting the forms of exploitation that it facilitates.

\section{Data}\label{sec:data}
This research uses three datasets derived from a database of 240 million ASW ads collected between June 2020 and February 2025. Each dataset serves a distinct purpose in our model development pipeline.

\subsection{Unique Posts Dataset}
The \textit{unique posts} dataset includes all unique post texts observed across the ASW ads in our database. In total, this dataset contains 19,833,258 post texts. This dataset serves as the corpus for all pre-training tasks.

\subsection{Triplet Dataset}
The \textit{triplet dataset} includes 4 million (anchor, positive, negative) tuples designed for fine-tuning sentence transformer models. The texts composing each tuple are selected such that the anchor and positive texts are more similar than the anchor and negative texts. 

To generate this dataset, we first construct a graph representation of our 240 million ASW ads using the associated post text and perceptual hash values for images (see \citet{keskin2021cracking} for additional information on this approach). This graph contains 107,861,111 vertices (19,833,258 unique post texts and 88,027,853 perceptual hash values) that are connected by 255,369,358 edges. Next, we extract all connected components and assume the data associated with each of the 1,883,588 non-giant components can be attributed to an individual or unique posting entity. We randomly select text pairs from the same non-giant component to obtain anchor and positive examples, and we randomly choose text from a different non-giant component to obtain negative examples. When selecting negative examples, we employ a \textit{hard negatives} approach that only allows selections with a cosine similarity greater than 0.2 based on comparing the text's TF-IDF encoding to that for the anchor text. The use of hard negatives ensures that negative examples exhibit some similarity to the anchor, which forces the model to learn more complex patterns rather than just detecting common keywords.

\subsection{Classification Dataset}
The \textit{classification} dataset includes 1 million text pairs constructed from the extracted graph components. Specifically, 200,000 pairs are selected from the same component, and 800,000 are chosen from different components. We create the dataset in an imbalanced fashion to ensure that trained models are sensitive to such imbalances, which our intuition suggests are likely to occur. Each pair of texts is associated with a binary label that indicates whether the texts originate from the same component (1) or different components (0). We use this dataset to evaluate the efficacy of binary classifiers for authorship verification trained using various sparse and dense encoding methods.

\section{Models}
Previous research involving AWS data text analysis has used both sparse and dense text representations. Thus, we also consider both sparse and dense representations, independently and combined as ensembles, for the author verification task. This section describes the sparse and dense models we consider for converting ASW data into numerical vector representations.

\subsection{Sparse Models}\label{subsec:sparse_models}
The sparse representations we consider are based on Term Frequency-Inverse Document Frequency (TF-IDF), a numerical metric designed to approximate word importance in documents comprising a corpus. \citet{sparck1972statistical} introduces the TF-IDF metric, which is comprised of two components:

\begin{itemize}
    \item Term Frequency (TF) - The normalized frequency of term $t$ in document $d$ is calculated as:
    $$\text{TF}(t,d) = \frac{\text{occurrences of term } t \text{ in document } d}    {\text{total number of terms in document } d}.$$
    \item Inverse Document Frequency (IDF) - A measure of how important a term $t$ is across the entire corpus:
    $$ \text{IDF}(t) = \log\left(\frac{N}{n_{t}}\right),$$
    where $N$ is the total number of documents in the corpus, and $n_t$ is the number of documents containing term $t$.
\end{itemize}

\noindent The TF-IDF score for a word $t$ relative to document $d$ is computed as:
$$\text{TFIDF}(t,d) = \text{TF}(t,d) \times \text{IDF}(t).$$

\noindent The logarithm in the IDF term reduces the impact of common terms, limits the range of the document frequency distribution, and reduces the distribution's skewness. These characteristics highlight rare terms that can effectively distinguish documents. TF-IDF's advantages are that it is: ($i$) intuitive and easy to implement, ($ii$) effective at identifying distinctive terms in a document, and ($iii$) can outperform more complex methods on tasks where matching keywords is essential. However, TF-IDF ($i$) can be biased when documents have variable lengths, ($ii$) does not capture the meaning of words or context, and ($iii$) is sensitive to misspellings.

TF-IDF requires a method to convert text into a list of tokens. In addition to considering WordPiece tokenizers, as are used for \texttt{BERT}, we also consider a custom method designed to handle emojis observed in ASW ad texts. Listing \ref{lst:tokenization} provides Python code for a function that applies the custom tokenization method to a single piece of text. 

\begin{lstlisting}[
caption={Tokenization Method}, 
label={lst:tokenization}, 
frame=single,
language=Python,
basicstyle=\tiny
]
def tokenize_with_emojis(
    text: str,
) -> list[str]:
    import re
    import string

    # This pattern matches emoji characters
    emoji_pattern = re.compile(
        r'['
        # Basic Emoji
        r'\U0001F600-\U0001F64F'  # Emoticons
        r'\U0001F300-\U0001F5FF'  # Symbols & Pictographs
        r'\U0001F680-\U0001F6FF'  # Transport & Map Symbols
        r'\U0001F700-\U0001F77F'  # Alchemical Symbols
        r'\U0001F780-\U0001F7FF'  # Geometric Shapes Extended
        r'\U0001F800-\U0001F8FF'  # Supplemental Arrows-C
        r'\U0001F900-\U0001F9FF'  # Supplemental Symbols & Pictographs
        r'\U0001FA00-\U0001FA6F'  # Chess Symbols
        r'\U0001FA70-\U0001FAFF'  # Symbols & Pictographs Extended-A
        r'\U0001FB00-\U0001FBFF'  # Symbols & Pictographs Extended-B
        r'\U0001FC00-\U0001FCFF'  # Symbols & Pictographs Extended-C
        r'\U0001FD00-\U0001FDFF'  # Symbols & Pictographs Extended-D
        r'\U0001FE00-\U0001FEFF'  # Symbols & Pictographs Extended-E
        r'\U0001FF00-\U0001FFFF'  # Symbols & Pictographs Extended-F
        
        # Additional Symbol Blocks
        r'\U00002600-\U000026FF'  # Miscellaneous Symbols
        r'\U00002700-\U000027BF'  # Dingbats
        r'\U0000FE00-\U0000FE0F'  # Variation Selectors
        r'\U0001F000-\U0001F02F'  # Mahjong Tiles
        r'\U0001F0A0-\U0001F0FF'  # Playing Cards
        r'\U0001F100-\U0001F1FF'  # Enclosed Alphanumeric Supplement (including flags)
        r'\U0001F200-\U0001F2FF'  # Enclosed Ideographic Supplement
        r'\U0001F300-\U0001F5FF'  # Miscellaneous Symbols and Pictographs
        r'\U0001F170-\U0001F251'  # Enclosed Alphanumerics
        
        # Symbols commonly used as or with emojis
        r'\U00002000-\U0000206F'  # General Punctuation
        r'\U00002100-\U00002BFF'  # Letterlike Symbols, Number Forms, Arrows, Math Operators, etc.
        r'\U00003000-\U0000303F'  # CJK Symbols & Punctuation
        
        # Modifier and ZWJ sequences
        r'\U0000200D'            # Zero Width Joiner
        r'\U0000FE0F'            # Variation Selector-16 (for emoji presentation)
        r'\U0001F3FB-\U0001F3FF' # Skin tone modifiers
        
        # Regional Indicator Symbols (used for flag emojis)
        r'\U0001F1E0-\U0001F1FF' # Regional indicator symbols
        r']',
        flags=re.UNICODE
    )

    # First, add spaces around emojis
    text = ''.join([char for char in text.lower() if char not in string.punctuation])
    text = emoji_pattern.sub(r' \g<0> ', text)

    # Split the text into tokens
    tokens = text.split()

    # Remove any empty tokens
    tokens = [token for token in tokens if token.strip()]

    return tokens
\end{lstlisting}

\begin{figure}[htb]
    \centering
    \includegraphics[width=0.75\linewidth]{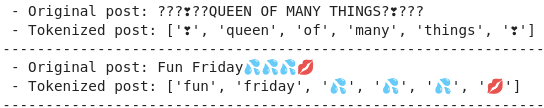}
    \caption{Tokenization Examples}
    \label{fig:tokenization-examples}
\end{figure}

\noindent Figure \ref{fig:tokenization-examples} uses two examples to demonstrate how our custom tokenization method processes texts containing emojis and punctuation. 

The following list specifies four sparse models we consider. Note that three models use custom BertWordPiece tokenizers with vocabularies of different sizes. We train these BertWordPiece tokenizers once, using the Python \texttt{tokenizers} library, and use them for the different variations of sparse and dense models we consider. We use the default TF-IDF implementation from \texttt{scikit-learn} \citep{scikit-learn}.

\begin{itemize}
    \item \texttt{TFIDF} - TF-IDF with tokenization performed by algorithm given in Listing \ref{lst:tokenization}
    \item \texttt{TFIDF-New15261} - TF-IDF with custom 15,261 vocabulary BertWordPiece tokenizer
    \item \texttt{TFIDF-New30522} - TF-IDF with custom 30,522 vocabulary BertWordPiece tokenizer
    \item \texttt{TFIDF-New45783} - TF-IDF with custom 45,783 vocabulary BertWordPiece tokenizer
\end{itemize}

\subsection{Dense Models}
Encoder-only foundation models like \texttt{BERT} generate contextualized token representations. When used directly for classification tasks, these models typically encode contextual information in special token embeddings (e.g., the classification \texttt{[CLS]} and separator [SEP] tokens in \texttt{BERT}). However, for similarity-based tasks, \textit{sentence transformers} enhance performance by adding a pooling layer over token embeddings to generate a sentence-level vector representation. These models are fine-tuned using a \textit{contrastive learning} task, where data triplets are used to produce \textit{similarity-aware embeddings} \citep{reimers2019sentence}. Ultimately, we use simple classification methods to convert similarity scores from such sentence transformer embeddings to author verification predictions. Thus, all of our dense models are used to yield a fine-tuned sentence transformer. 

We explore three pathways to arrive at sentence transformer models. The first and most straightforward pathway is to use a pre-trained sentence transformer model available on \href{https://huggingface.co/}{Hugging Face}. The second pathway is to use a pre-trained foundation model and fine-tune the model using contrastive learning to produce a sentence transformer model. The third pathway is to pre-train a custom foundation model that is then fine-tuned using contrastive learning to generate the sentence transformer model. The following subsections describe the specific variants we consider for each pathway.

\subsubsection{Pathway 1: Pre-trained Sentence Transformers}
We consider the following pre-trained sentence transformers in this study. When using one of these methods, we only need to download the model and encode the text to generate a vector representation.

\begin{itemize}
    \item \texttt{all-MiniLM-L6-v2}: A compact and efficient model that performs well on benchmarks for similarity tasks (\url{https://huggingface.co/sentence-transformers/all-MiniLM-L6-v2})
    \item \texttt{mxbai-embed-large-v1}: A larger model with improved performance on semantic similarity tasks \url{https://huggingface.co/mixedbread-ai/mxbai-embed-large-v1}
\end{itemize}

\subsubsection{Pathway 2: Pre-trained Foundation Models}
We consider the following pre-trained foundation models in this study. When using one of these methods, we first fine-tune the model using the triplet dataset with a contrastive learning task to produce a sentence transformer that can encode ASW ad texts.

\begin{itemize}
    \item \texttt{bert-base-uncased}: \url{https://huggingface.co/google-bert/bert-base-uncased}
    \item \texttt{RoBERTa-base}: \url{https://huggingface.co/FacebookAI/roberta-base}
    \item \texttt{ModernBERT-base}: \url{https://huggingface.co/answerdotai/ModernBERT-base}
\end{itemize}

\subsubsection{Pathway 3: Custom Foundation Models}
We pre-train custom foundation models from the \texttt{bert-base-uncased} configuration that is available in the \texttt{transformers} Python library. These models require pre-training on our unique posts dataset to learn vector representations for tokens that compose the vocabulary. A mean pooling layer is added to the pre-trained model, and it is fine-tuned on the triplet dataset with a contrastive learning task to produce a sentence transformer. These models offer the most flexibility in handling the unique characteristics of ASW text, but they require intensive computation. 

As was done for the TF-IDF variants presented in Section \ref{subsec:sparse_models}, we experiment with vocabulary sizes that are smaller and larger than the original \texttt{BERT} implementation. We also consider two different learning tasks during pre-training: ($i$) \textit{masked language modeling} (MLM) and ($ii$) \textit{whole word masked language modeling} (WWMLM). MLM is used in the original \texttt{BERT} implementation and involves randomly masking individual tokens (including subword pieces) that the model learns to predict based on surrounding tokens. In contrast to MLM, WWMLM masks entire words rather than subword tokens. Our motivation for considering WWMLM is that it forces the model to understand complete semantic units, rather than reducing the set of possible choices for a prediction based on common subword occurrences.

In contrast to the other pathways, this approach yields an entirely custom model with the most flexibility in representing the atypical text patterns observed in ASW data. However, it also requires the most intensive computation. One additional thing to note is that in addition to experimenting with vocabulary sizes that differ from the original \texttt{BERT} implementation, we consider variants that use a \textit{whole word masked language modeling} task in addition to the standard \textit{masked language modeling}. For both modeling tasks, we employ dynamic masking during pre-training as was done for \texttt{RoBERTa} because it is the default masking approach when training models with the \texttt{transformers} library.

We evaluate six custom model variants with different vocabulary sizes and pre-training tasks. All of our variants are based on the \texttt{bert-base-uncased} model from HuggingFace. The six variants are:

\begin{itemize}
    \item \texttt{BERT-New15261-MLM}: 15,261 vocabulary BertWordPiece tokenizer with MLM
    \item \texttt{BERT-New30522-MLM}: 30,522 vocabulary BertWordPiece tokenizer with MLM
    \item \texttt{BERT-New45783-MLM}: 45,783 vocabulary BertWordPiece tokenizer with MLM
    \item \texttt{BERT-New15261-WWMLM}: 15,261 vocabulary BertWordPiece tokenizer with WWMLM
    \item \texttt{BERT-New30522-WWMLM}: 30,522 vocabulary BertWordPiece tokenizer with WWMLM
    \item \texttt{BERT-New45783-WWMLM}: 45,783 vocabulary BertWordPiece tokenizer with WWMLM
\end{itemize}

\subsection{Training Optimizations}
We use the same core implementation details as the original \texttt{BERT} variants \citep{devlin-etal-2019-bert} with a couple of optimizations to address hardware constraints observed during initial experimentation. All computation was performed on a single NVIDIA RTX 6000 Ada GPU with 18,176 CUDA cores and 48 GB of RAM.

\begin{figure}[htb]
    \centering
    \includegraphics[width=\linewidth]{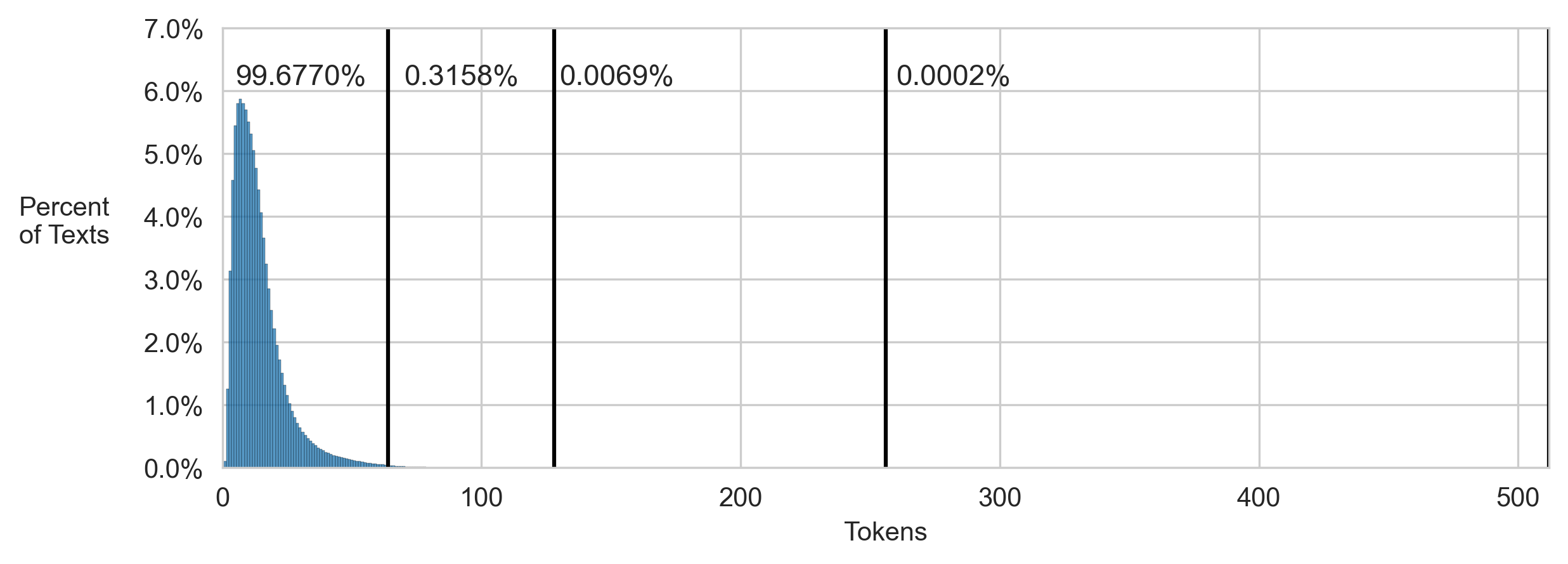}
    \caption{Text Length Distribution - BertWordPiece Tokenization with 30,522 Vocabulary}
    \label{fig:BertWordPiece-30522-text-lengths}
\end{figure}

One observation we made during initial tests was that the default \texttt{BERT} sequence length of 512 tokens required a relatively small batch size of 32 to avoid GPU memory issues, which reduced training efficiency and hindered model convergence. To address this, we analyzed the distribution of tokenized post lengths in our unique posts dataset. This analysis revealed that most of our tokenized posts fell well below the maximum 512-sequence length. For example, when using the 30,522 token vocabulary, we found that 99.677\% of the post texts include at most 64 tokens, and 99.9928\% include at most 128. Figure \ref{fig:BertWordPiece-30522-text-lengths} shows a histogram of the tokenized post length using this tokenizer that includes black horizontal reference lines at positions 64, 128, 256, and 512 on the horizontal axis.

Given that transformer self-attention mechanisms have a computational complexity of $O(n^2d)$, where $n$ is sequence length and $d$ is hidden dimension size, reducing the sequence length enables larger batch sizes and faster training. Based on these findings, we experiment with maximum sequence lengths of 64 and 128 (instead of the original 512) for each custom pre-trained variant, resulting in 12 total model variants (i.e., 6 models × 2 sequence lengths). 

Another optimization we employ is the use of \textit{Brain Floating Point 16-bit} (bfloat16) mixed precision for training and validation \citep{kalamkar2019study}. The bfloat16 format was developed primarily for accelerating machine learning workloads. We use the \texttt{pytorch} library for all training. By default, \texttt{pytorch} \textit{tensors} utilize 32-bit floating point (FP32) precision. Using bfloat16 requires half the memory to represent tensors, resulting in significant performance improvements. However, bfloat16 has the same dynamic range as FP32, allowing it to avoid numerical instability in deep neural network models.

\section{Experiment 1 - Authorship Verification Effects}\label{sec:experiment1}
\begin{figure}[htbp]
    \centering
    \includegraphics[width=\linewidth]{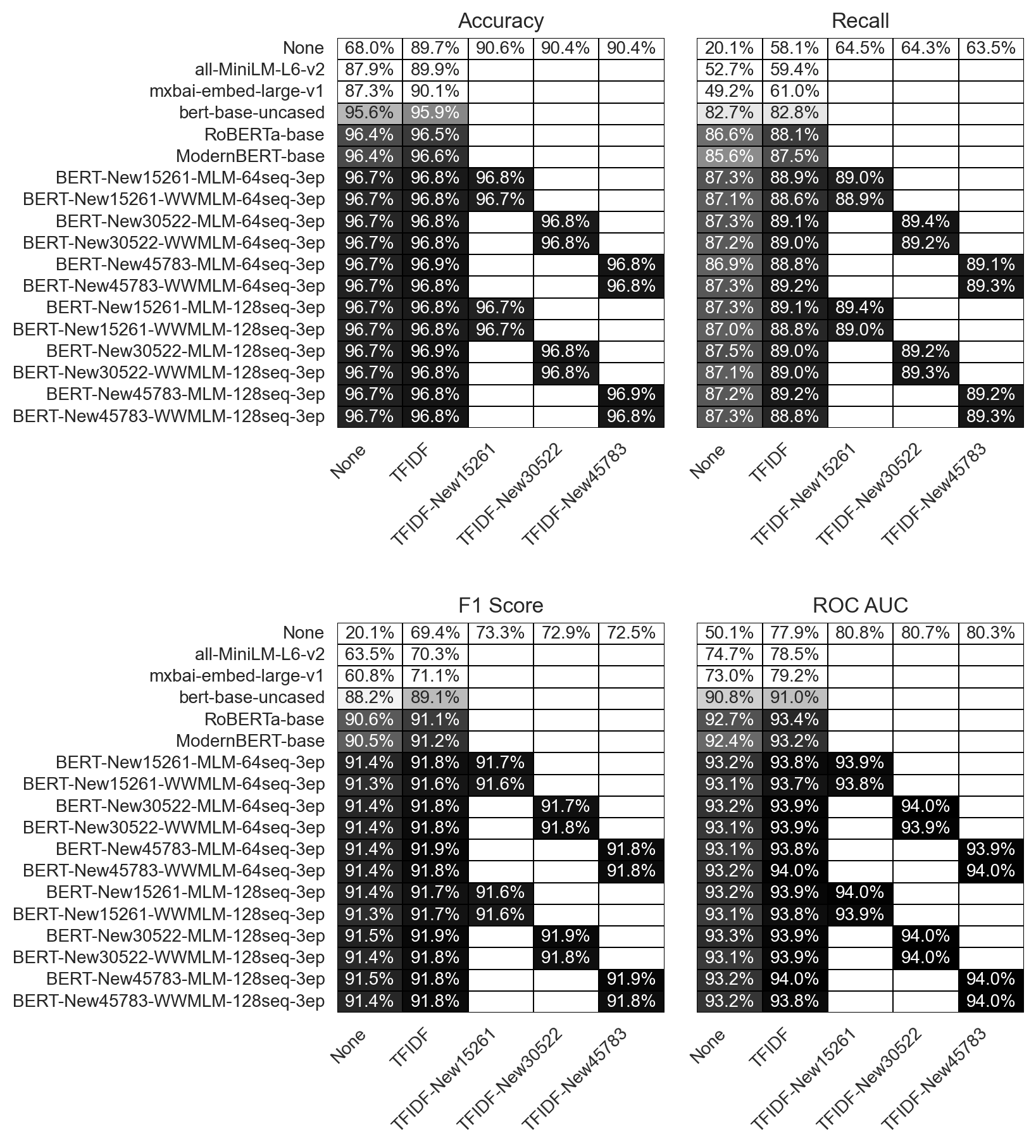}
    \caption{Classification Results}
    \label{fig:classification_results}
\end{figure}

This section presents the first of two sets of experiments, which examines how the various sparse and dense encoding methods affect the performance of an authorship verification task based on our classification dataset. For this initial experiment, we perform limited pre-training of three epochs for all custom pre-trained variants. We evaluate all sparse and dense models independently in ensembles of sparse and dense models. We maintain tokenizer consistency for ensemble configurations, pairing sparse models with dense models that use the same tokenizer. The only exceptions are cases that utilize TF-IDF with the custom tokenization method presented in Listing \ref{lst:tokenization}. We include a random baseline that guesses whether two posts are likely to be attributed to the same author with a positive likelihood of 20\%, matching the positive instance proportion in our classification dataset. This random baseline is represented by sparse and dense models of \texttt{None}. Figure \ref{fig:classification_results} uses heatmaps to display the accuracy, recall, F1 score, and ROC AUC for all tested configurations, with sparse models varying along the horizontal axes and dense models along the vertical axis.

We make several observations regarding Figure \ref{fig:classification_results}:
\begin{itemize}
    \item Sparse Model Performance: The simple TF-IDF method achieves surprisingly high predictive accuracy when used alone. BertWordPiece tokenization improves all metrics for TF-IDF models, though recall is a weakness when compared to dense models.
    \item Pre-trained Sentence Transformers: The general-purpose pre-trained sentence transformers (\texttt{all-MiniLM-L6-v2} and \texttt{mxbai-embed-large-v1}) perform poorly, being outperformed by standalone sparse models. Their performance improves when combined with sparse models in ensembles, but still lags behind other dense model alternatives.
    \item Fine-tuned Models: Fine-tuning sentence transformers on our data yields substantial improvements. Both \texttt{RoBERTa}- and \texttt{ModernBERT}-based models consistently outperform \texttt{bert-base-uncased}, highlighting the importance of selecting a base model and the enhancements made by these two variants.
    \item Custom Pre-trained Models: Our domain-specific pre-trained models outperform all other approaches, whether used independently or in ensembles. These models show particularly significant improvements in recall, the metric where other approaches struggle most.
\end{itemize}

\begin{figure}[htb]
    \centering
    \includegraphics[width=\linewidth]{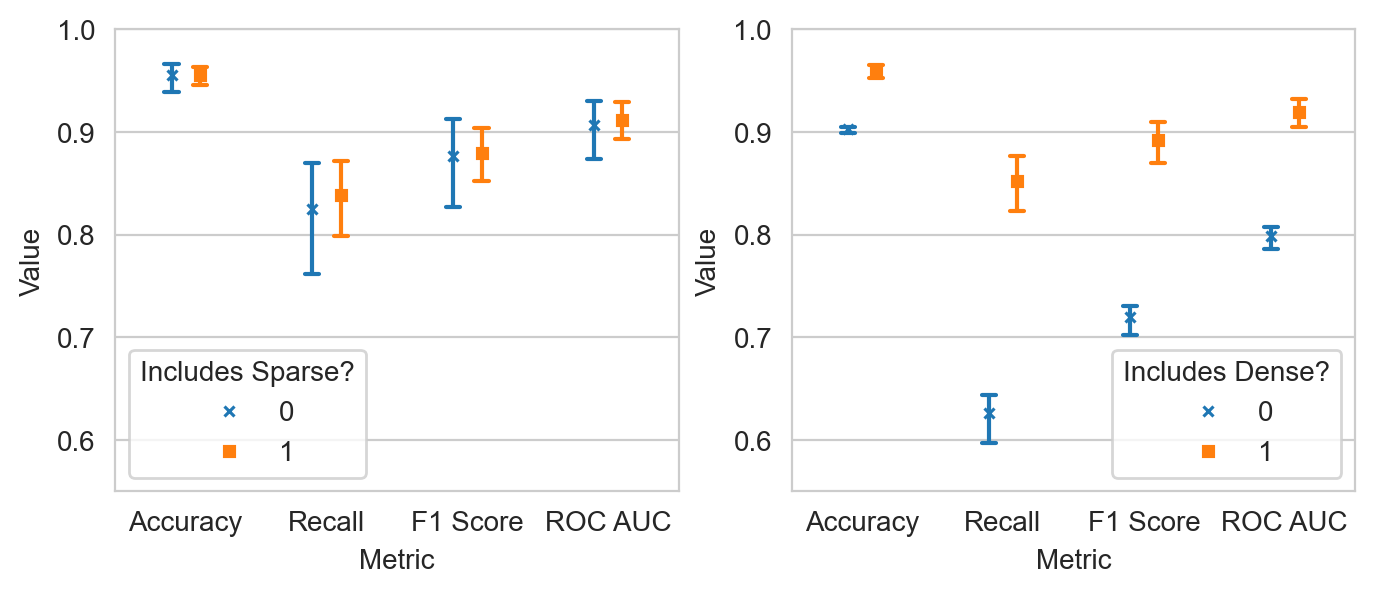}
    \caption{Main Effects of Including Sparse and Dense Models}
    \label{fig:sparse_dense_inclusion_effects}
\end{figure}

We analyzed our experimental results to assess the effects of vocabulary size, modeling task (MLM vs. WWMLM), and sequence length. These analyses identified no statistically significant impact on classification performance. However, the inclusion of sparse or dense embeddings did have a significant effect on the results.  Figure \ref{fig:sparse_dense_inclusion_effects} uses 95\% confidence intervals to show these effects for all performance measures. Specifically, the left subplot shows confidence intervals for including sparse embeddings during classifier training, and the right subplot shows the impact of including dense embeddings.

\begin{figure}[htb]
    \centering
    \includegraphics[width=\linewidth]{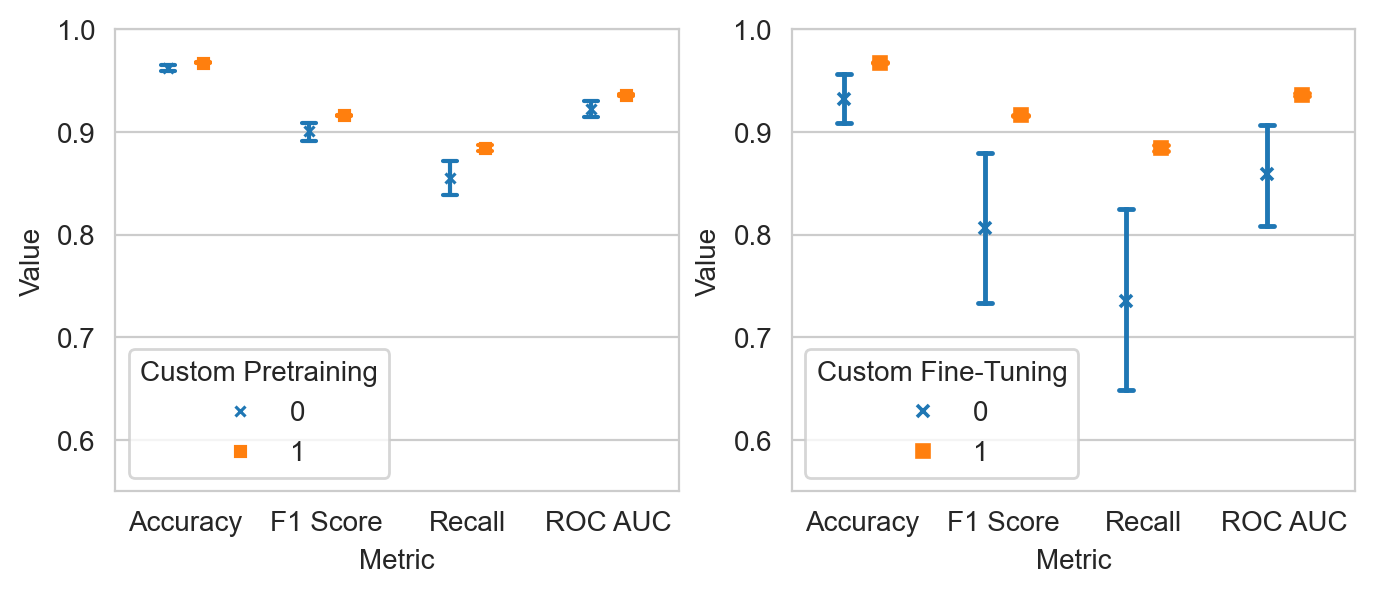}
    \caption{Main Effects of Custom Pre-training and Fine-tuning}
    \label{fig:custom_pretraining_finetuning_effects}
\end{figure}

The previous figure demonstrates that including dense embeddings substantially improves all metrics, with recall showing the most dramatic improvement. To further understand the benefits of our approach, Figure \ref{fig:custom_pretraining_finetuning_effects} presents 95\% confidence depicting the value of custom pre-training (left) and fine-tuning (right).

While the impact of custom pre-training is less dramatic, it still produces statistically significant improvements across all metrics, particularly for recall. Custom fine-tuning shows more substantial effects, with noticeable improvements across all metrics.

This experiment demonstrates that pre-training custom foundation models and fine-tuning them on ASW-specific datasets significantly benefits authorship verification performance. Even with minimal pre-training (three epochs), our custom models outperform state-of-the-art models like \texttt{RoBERTa} and \texttt{ModernBERT}. Thus, domain-specific adaptation of foundation models is beneficial for specialized text analysis tasks, such as authorship verification based on ASW data. However, given the limited number of pre-training epochs we conduct, our models may be undertrained, potentially leaving additional performance improvements unrealized. Our second experiment examines the marginal benefits of extended pre-training.

\section{Experiment 2 - Extended Pre-Training}

The previous experiment established the value of pre-training custom foundation models for ASW text analysis. However, the pre-training durations were purposefully limited to enable efficient exploration of key modeling choices, e.g., vocabulary size and modeling task. Our second experiment examines the marginal benefits of extended pre-training on authorship verification performance.

\begin{figure}[htb]
    \centering
    \includegraphics[width=\linewidth]{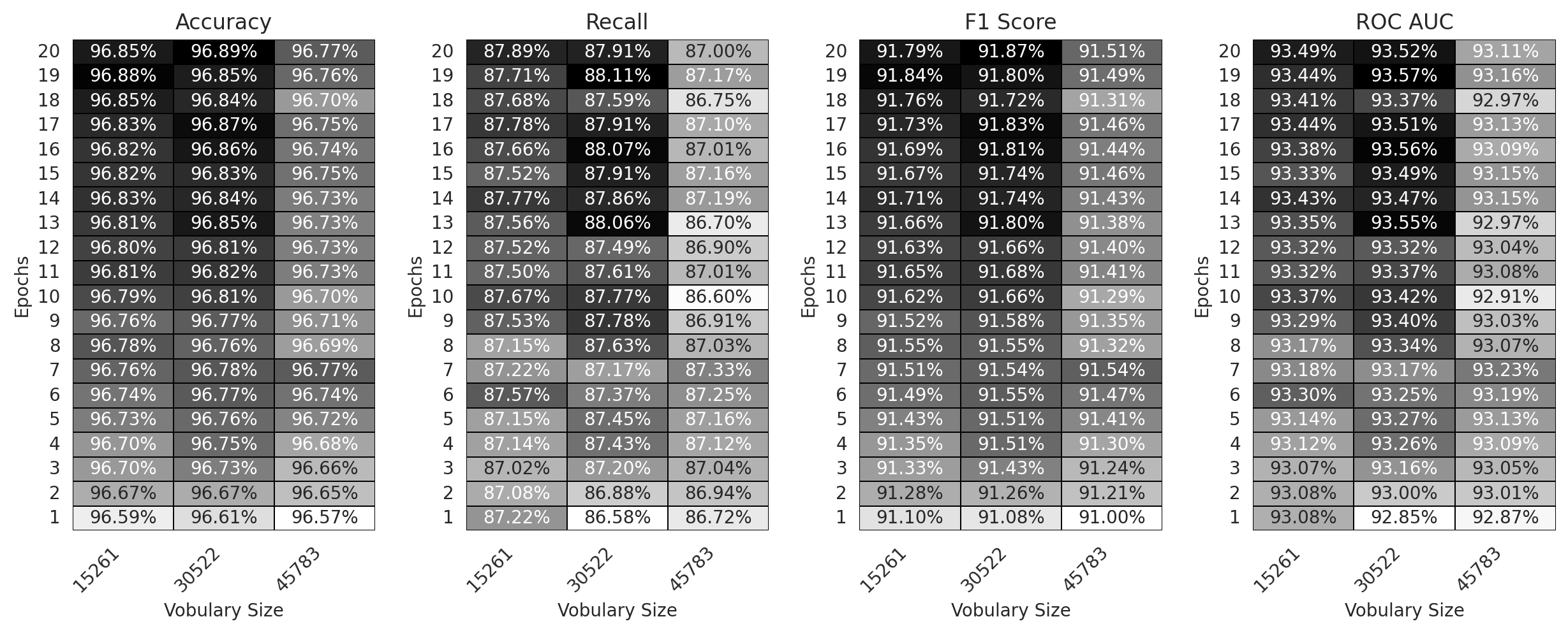}
    \caption{Effects of Extended Pre-training}
    \label{fig:long_classification_results}
\end{figure}

Based on our initial findings, we selected the most promising model configurations for extended training. Specifically, we consider the \texttt{BERT-New15261-MLM}, \texttt{BERT-New30522-MLM}, and \texttt{BERT-New45783-MLM} variants with a sequence length of 64 tokens and 20 training epochs. For each model variant, we saved checkpoints after each training epoch and fine-tuned a separate sentence transformer model from each checkpoint. This produced 60 sentence transformer models (3 vocabulary sizes × 20 epochs), which we evaluated on our classification dataset. We did encounter issues with exploding gradients for the \texttt{BERT-New45783-MLM} variant. We were using gradient clipping, but still encountered issues. Thus, we were forced to reduce the learning rate from $1 \times 10^{-4}$ to $1 \times 10^{-5}$, which we acknowledge influenced its performance. Figure \ref{fig:long_classification_results} uses heatmaps to show the experiment results for the four performance measures. In each heatmap, the number of epochs for the pre-trained model increases along the vertical axis, ranging from 1 at the bottom to 20 at the top. The vocabulary size varies along the horizontal axis.

The extended pre-training experiment reveals several important findings. First, we realize performance improvements for all metrics with additional pre-training epochs, confirming that our initial three-epoch models were undertrained. Second, performance gains begin to plateau relatively early (i.e., after 3 - 5 epochs), with marginal improvements and some variability beyond this point. Third, the high levels of performance are achieved using only dense representations without requiring ensembles with sparse representations. Fourth, the \texttt{BERT-New30522-MLM} variant, trained for 20 epochs, consistently achieves performance at or near peak levels across all metrics, making it our recommended model for downstream applications.

\section{Applications}

We present three applications to demonstrate the practical utility of our recommended model: the \texttt{BERT-New30522-MLM} variant pre-trained with a 64-token sequence length for 20 epochs and its corresponding fine-tuned sentence transformer. The applications we consider are:

\begin{itemize}
    \item GC Decomposition - decomposing the giant component that emerges in ASW data graph representations into smaller clusters based on authorship verification of connected texts,
    \item Semantic Search -efficiently retrieving ASW posts that are semantically similar, and
    \item Token Understanding -Analyzing context-specific meanings of tokens and emojis in ASW advertisements.
\end{itemize}

\subsection{Data}\label{sec:application_data}
Our applications use the open dataset described in \citet{Freeman-2025-multisite}, which provides a multi-site sample of ASW ad data. The dataset contains over 10 million advertisements collected from nine sites between May 1, 2022, and August 1, 2022. To protect privacy, the dataset obfuscates numerical data included in post texts, anonymizes URLs and site names, and provides perceptual hash values instead of images. Table \ref{tab:data_columns} provides descriptions for columns we use in our applications, where the descriptions are copied verbatim from \citet{Freeman-2025-multisite}.

\begin{table}[htb]
    \centering
    \begin{tabular}{lp{10.5cm}}
    \hline 
    Column Name & Description \\ \hline
    \texttt{url} & An integer representing the URL associated with the ad. \\
    \texttt{site} & An integer representing the site the data was collected from. \\
    \texttt{post\_masked} & The raw heading text associated with the ad. \textbf{Note}: Numbers that appear in the text are obfuscated by replacing them with the character `*'. \\
    \texttt{post\_int} & An integer representing the raw heading text associated with the ad. \textbf{Note}: All occurrences of a specific text are assigned to the same integer value. \\
    \texttt{phone\_int} & An integer representing the phone number associated with the ad. Missing values are represented as \texttt{None}. \\
    \texttt{phash16} & The perceptual hash for a single image associated with the ad, computed using the \texttt{imagehash} library. \\ \hline
    \end{tabular}
    \caption{Relevant Data Columns}
    \label{tab:data_columns}
\end{table}

\subsection{GC Decomposition}
This section demonstrates the ability of the devised approach to assist in the decomposition of the giant component that forms in graph representations of ASW ad data. For each of the nine sites in the dataset, we:
\begin{enumerate}
    \item construct a bipartite graph representation with: ($i$) vertices representing unique \texttt{post\_int} and \texttt{phash16} values, and ($ii$) edges connecting vertices that appear together in at least one advertisement,
    \item extract the giant component from the graph representation and project it to include only post-to-post connections,
    \item compute cosine similarity scores between connected posts using embeddings generated by our fine-tuned sentence transformer,
    \item filter edges based on cosine similarity thresholds ranging from 0.05 to 0.95, and
    \item extract the remaining connected components after edge filtering.
\end{enumerate}
\noindent We use the \textit{igraph} Python package for all graph construction and projection tasks. 

\begin{figure}[htb]
    \centering
    \includegraphics[width=\linewidth]{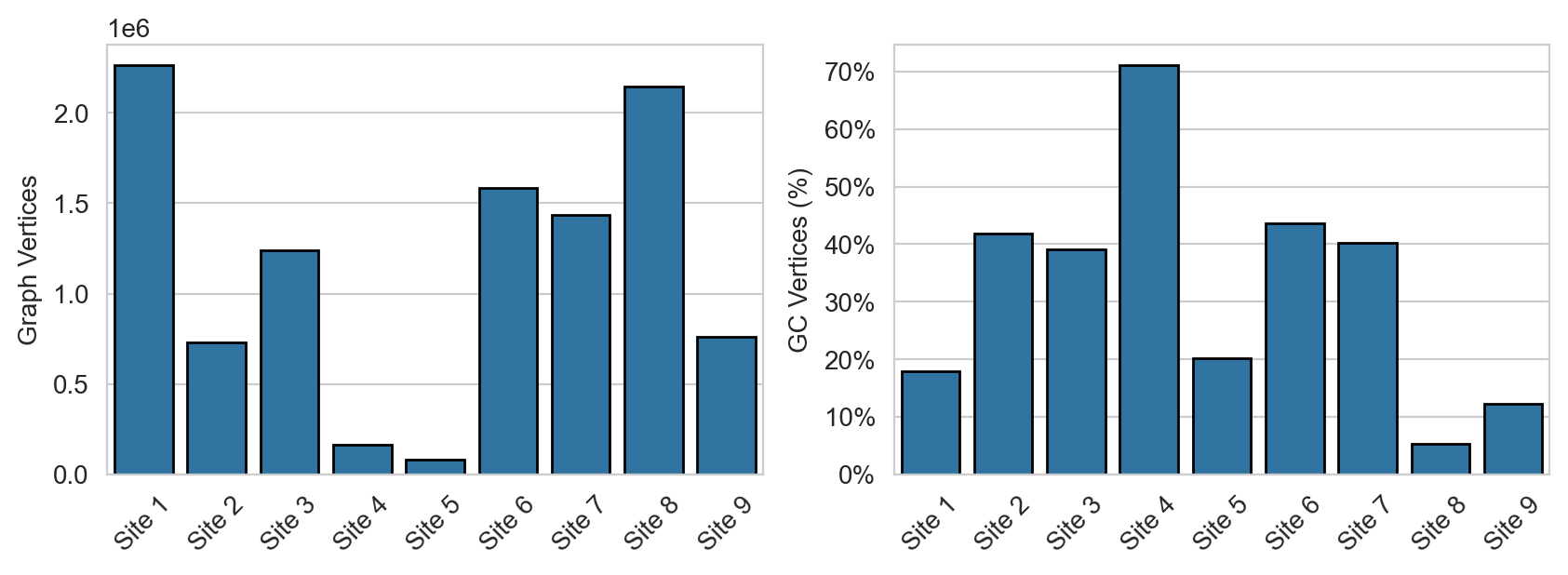}
    \caption{Graph and GC Size by Site}
    \label{fig:gc_stats}
\end{figure}

Figure \ref{fig:gc_stats} uses bar plots to show the size, i.e., number of vertices, of the graph representation for each site (left subplot) and the size of the associated GC, represented as a percent of the vertices in the graph representation. As the figure shows, the sites represented in the data vary significantly for these two measures.

\begin{figure}[htb]
    \centering
    \includegraphics[width=\linewidth]{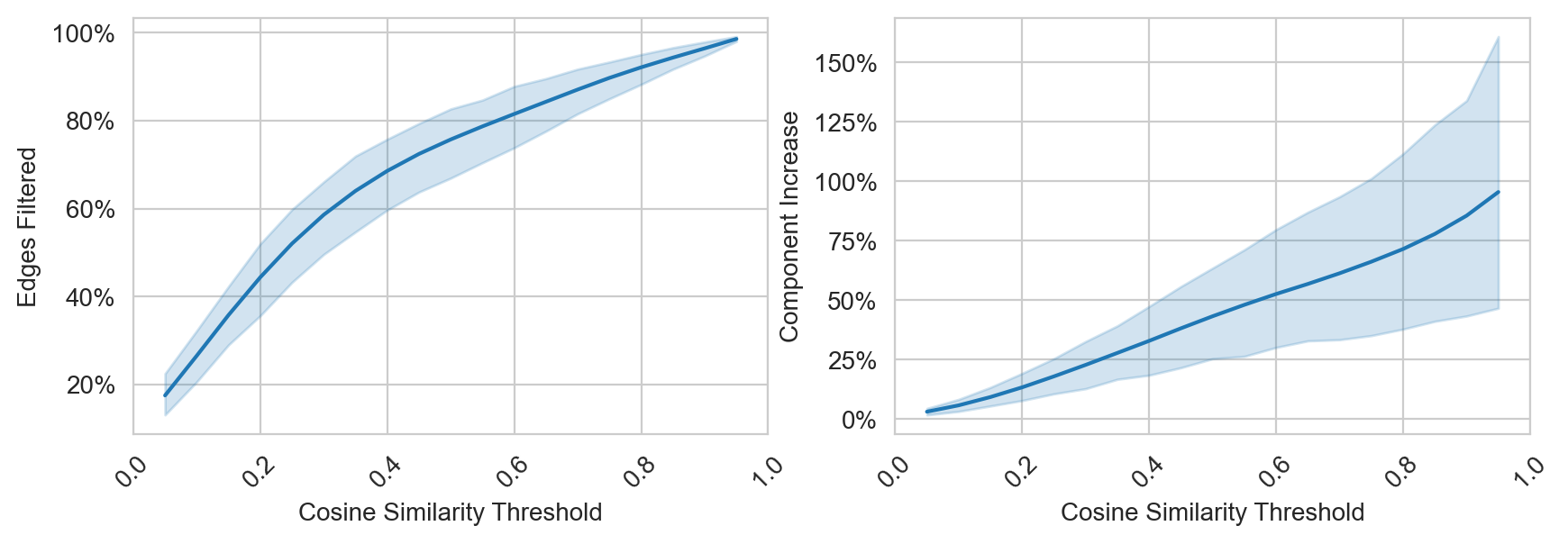}
    \caption{GC Filtering Effects}
    \label{fig:gc_filtering_effects}
\end{figure}

Figure \ref{fig:gc_filtering_effects} depicts the results of this filtering experiment. Specifically, the left subplot shows the percentage of edges filtered from the projected giant component representation as the cosine similarity threshold varies. The right subplot shows the increase in components that results from filtering, expressed as a percentage of the original graph representation's components. The light-shaded regions around each line represent the variance across sites. The results show that our model-based filtering approach effectively dismantles the observed giant components into smaller, more coherent groups. Moreover, the similarity threshold provides an intuitive parameter that practitioners can adjust based on their preferences.

\subsection{Semantic Search}
The emergence of dense methods for representing text has spurred significant innovation in semantic search and document retrieval. A typical process involves:
\begin{itemize}
     \item encoding documents using a dense vector representation,
     \item storing document embeddings in specialized vector databases or vector stores that use algorithms like Approximate Nearest Neighbor (ANN) search to all efficient similarity search in high-dimensional spaces,
     \item encoding queries using the same encoder as was used to generate document embeddings, and 
     \item retrieving the most similar documents from the vector database using metrics like cosine similarity or Euclidean distance.
\end{itemize}

\begin{figure}[htb]
    \centering
    \includegraphics[width=0.6\linewidth]{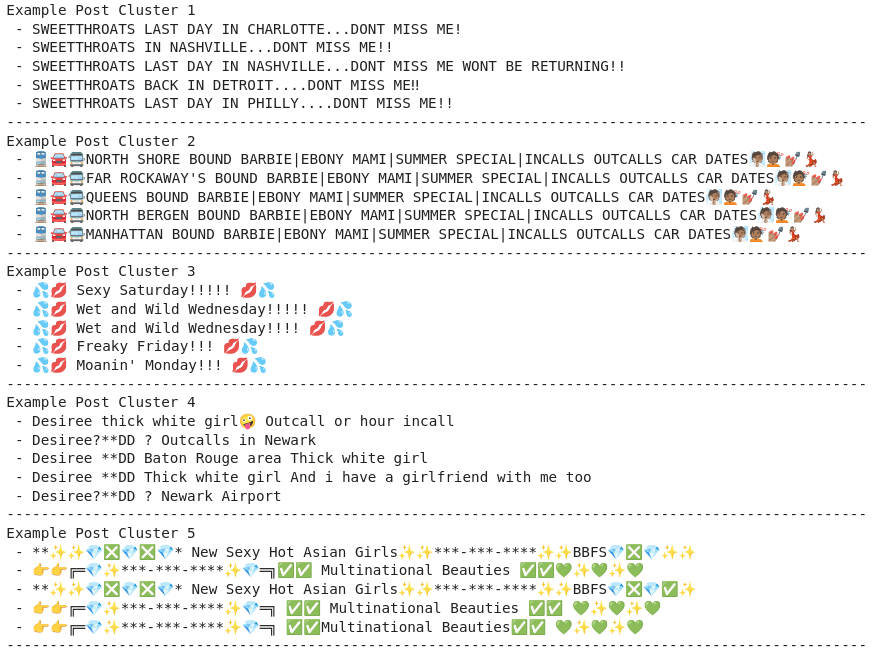}
    \caption{Post Cluster Examples}
    \label{fig:post_cluster_example}
\end{figure}

The dense representations generated by our custom sentence transformer enable such an approach for ASW data. To demonstrate this, we implement a simplified semantic search process that: ($i$) encodes all post texts using our recommended fine-tuned sentence transformer, ($ii$) projects the resulting embeddings to a lower-dimensional space using LocalMAP \citep{wang2024dimensionreductionlocallyadjusted}, ($iii$) applies HDBSCAN \citep{McInnes2017} to cluster the projected embeddings, and ($iv$) extracts clusters. Figure \ref{fig:post_cluster_example} shows example posts from a set of five selected clusters created using the described process.

Figure \ref{fig:post_cluster_example} shows the ability of the simple search process to identify highly similar posts. A key thing to note is that this clustering is achieved without a graph representation linking posts based on common ad information (e.g., common perceptual hashes). Admittedly, TF-IDF will also rate several of the clustered posts as similar due to common keywords. However, as we will see in the next section, our model has learned important information that allows it to assess similarity without common keywords.

\subsection{Token Understanding}
Our final application demonstrates how our custom foundation models enable insight regarding the lexicon used in ASW ad text. For this application, we do the following: ($i$) extract token embeddings from the embedding layer of our pre-trained model, ($ii$) map the embeddings to the corresponding tokens, and ($iii$) project the token embeddings into a two-dimensional representation using LocalMAP. 

\begin{figure}[htb]
    \centering
    \includegraphics[width=0.7\linewidth]{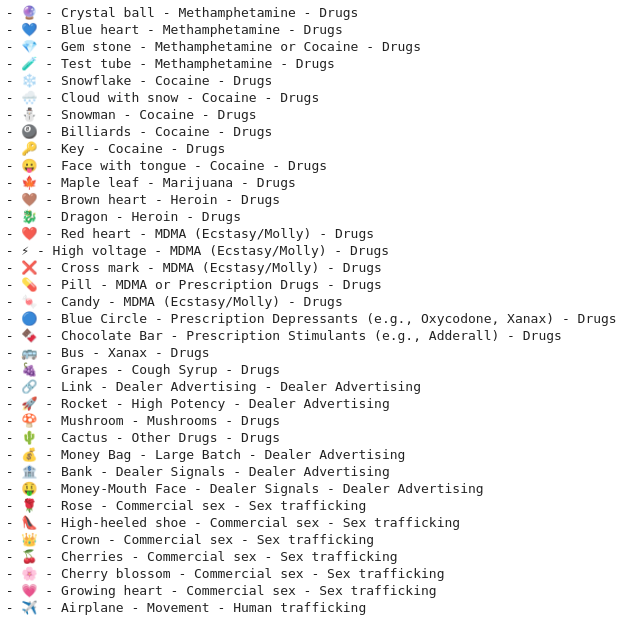}
    \caption{Selected \textit{Illicit} Emojis}
    \label{fig:emoji_meanings}
\end{figure}

First, we use the projected embeddings to understand the similarity among a set of  \textit{illicit emojis} depicted in Figure \ref{fig:emoji_meanings}. These emojis are reported to indicate commercial sex, human trafficking, or drug use \citep{dea2021emoji, darkowl2022emoji}. Each entry in the figure provides the emoji, its name, the suggested meaning, and the associated category, such as drugs.

\begin{figure}[htb]
    \centering
    \includegraphics[width=0.7\linewidth]{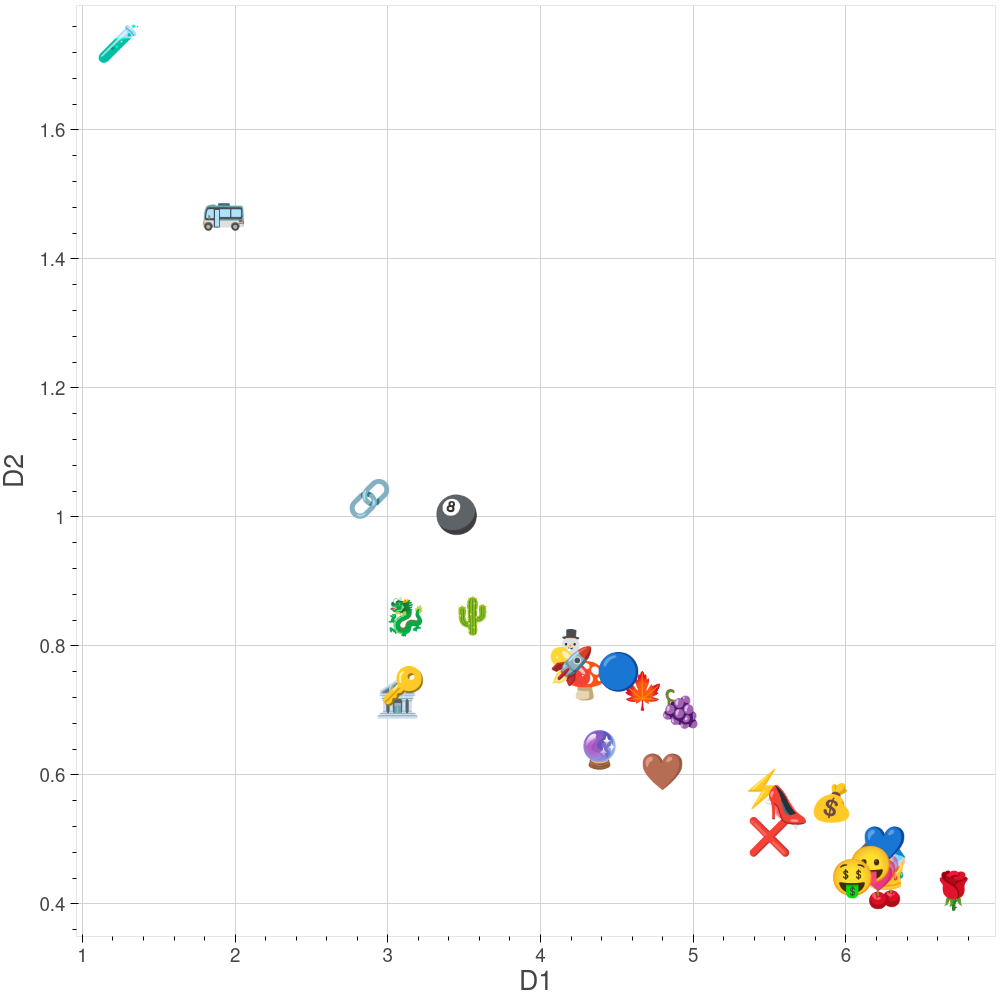}
    \caption{Illicit Emoji Relationships}
    \label{fig:illicit_emoji_scatterplot}
\end{figure}

Figure \ref{fig:illicit_emoji_scatterplot} uses a scatterplot to show the orientation of the matched illicit emojis in the projected embedding space. Note that we exclude instances where a token includes a selected emoji prefixed by something else, e.g., \#\#\textit{emoji}. As the figure shows, the projected embeddings cluster many of the suspected drug-, dealer-, and commercial sex-related emojis very close to one another. This clustering validates reports regarding emoji usage while providing additional nuance about their contextual relationships in ASW data.

\begin{figure}[htb]
    \centering
    \includegraphics[width=0.7\linewidth]{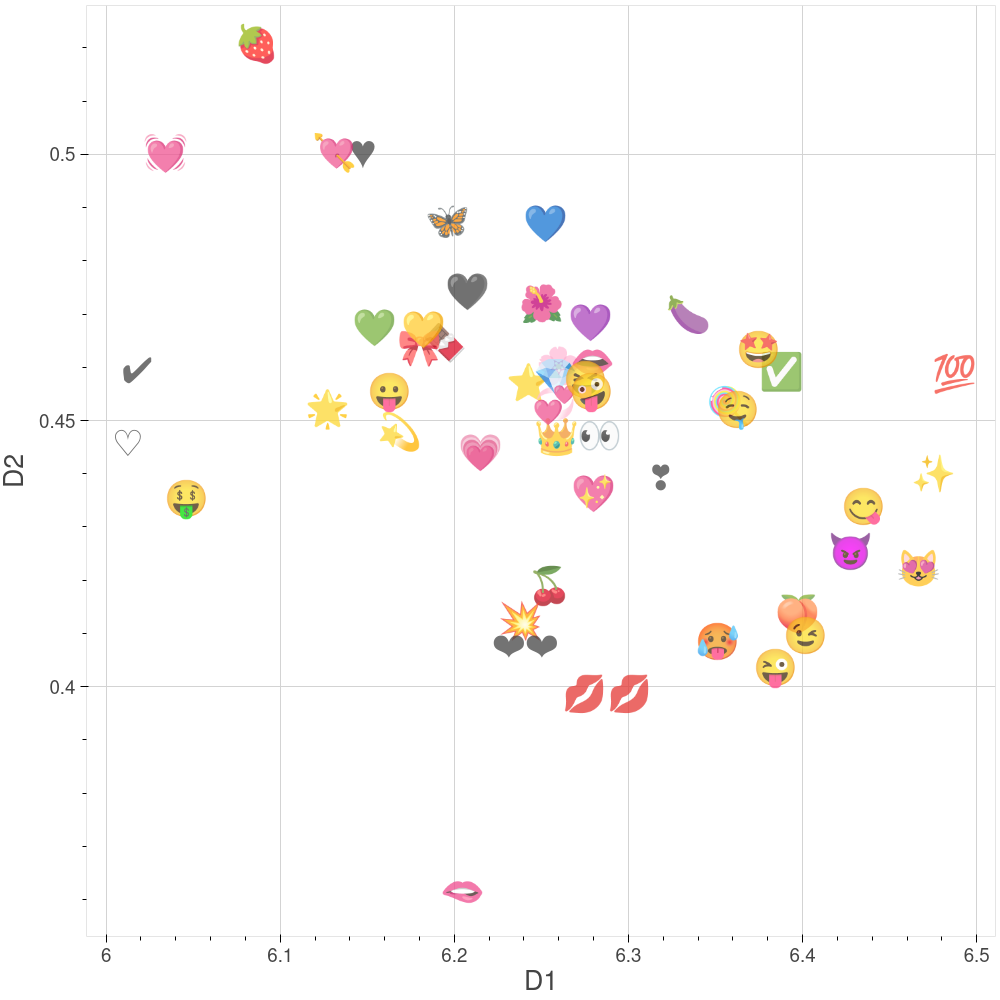}
    \caption{Crown Emoji Relationships}
    \label{fig:crown_emoji_scatterplot}
\end{figure}

To further demonstrate the model's contextual understanding, we examine the crown emoji, which reportedly indicates commercial sex/sex trafficking, and its 50 nearest neighbors in the projected embedding space. Figure \ref{fig:crown_emoji_scatterplot} uses a scatterplot to depict the position of these neighbors using a reduced scale for the horizontal and vertical axes to provide visual granularity. We see that several nearby emojis also appear in \ref{fig:emoji_meanings} as indicators of commercial sex or human trafficking activity. More importantly, we see additional emojis not on the list, indicating that the lexicon may be more expansive than believed or that it has evolved.

\section{Conclusion}
This research presents a comprehensive approach to analyzing text in commercial sex ads using custom language models, underscoring the significant value of employing these models for analyzing ASW text. By pre-training and fine-tuning models tailored to ASW data, the study achieves remarkable improvements in authorship verification tasks. These custom models, designed to understand the unique linguistic patterns and nuances of ASW ads, outperform existing state-of-the-art models, demonstrating superior capability in this specialized context.

One of the critical challenges highlighted in this research is the difficulty in obtaining the data used to train these models. The dataset we use to train the custom models, comprising over 240 million commercial sex ads collected over nearly five years, is not only vast but highly varied and inconsistent across sources. This makes the process of data collection, cleaning, and preparation particularly arduous. Despite these challenges, the custom models developed in this study show that with the right data and training, significant advancements can be made in transforming ASW data into actionable intelligence.

The practical applications of these models are profound. They enable more effective dismantling of erroneous links in graph representations of ASW data (giant component decomposition), enhance semantic search capabilities, and provide deeper insights into the use of language in these ads. These advancements have the potential to significantly aid in the fight against sex trafficking, offering law enforcement and other stakeholders powerful analytic tools to identify and disrupt sex trafficking networks. By leveraging custom language models, this research paves the way for more efficient and accurate analysis of ASW data, ultimately contributing to efforts to combat this serious issue and potentially save lives. Additionally, these models provide a unique tool for better understanding the linguistic nuances of ASW post texts and making connections to other illicit markets, such as drugs.

Despite our promising results, there are limitations. First, the sensitive nature of ASW data necessitates careful handling and privacy protections, which can limit data sharing and research reproducibility. The authors are happy to consider requests for data and our models. However, due to the sensitive nature of the data, we will consult with law enforcement partners we work with before providing any data or software. Second, while our models demonstrate impressive performance on a large and real dataset, monitoring and adaptation are necessary to keep up with evolving language patterns and obfuscation techniques. Finally, the computational resources required for model deployment are non-trivial, which can impede adoption by organizations with limited computing resources.

The development and application of computational tools for analyzing ASW data raise important ethical considerations that must be carefully addressed. First, the distinction between consensual sex work and sex trafficking is critical, and computational tools must be designed to avoid conflating these distinct phenomena. Our authorship verification approach focuses on confirming the common authorship of post text pairs and does not attempt to translate text into predictions regarding trafficking activity. Second, algorithm bias is a significant concern, as models trained on existing data may perpetuate or amplify existing biases in law enforcement attention. We recommend regularly evaluating model outputs to identify and mitigate such biases. Third, responsible deployment requires multidisciplinary collaboration among researchers, technologists, law enforcement, trafficking experts, and survivors to ensure that these tools support human-centered investigations rather than replacing critical human judgment. This research has been conducted with input from experts in human trafficking, digital privacy, law enforcement, and trafficking survivors to ensure that the capabilities were responsibly developed.

This research offers several promising areas for future research. First, our text-based models can be used in the development of multimodal models that incorporate both text and image data for more comprehensive ASW analysis. Second, our models can be used with spatiotemporal analysis techniques to detect evolving patterns in ASW text across space and time. Third, expanded token-level analyses can allow the development of more comprehensive lexicons of coded language used in ASW contexts. Finally, there is a need for developing intuitive, efficient, and interactive interfaces that allow domain experts to leverage these models without requiring technical expertise.






\begin{thebibliography}{26}
\expandafter\ifx\csname natexlab\endcsname\relax\def\natexlab#1{#1}\fi
\providecommand{\url}[1]{\texttt{#1}}
\providecommand{\path}[1]{#1}
\providecommand{\DOIprefix}{doi:}
\providecommand{\ArXivprefix}{arXiv:}
\providecommand{\URLprefix}{URL: }
\providecommand{\Pubmedprefix}{pmid:}
\providecommand{\doi}[1]{\href{http://dx.doi.org/#1}{\path{#1}}}
\providecommand{\Pubmed}[1]{\href{pmid:#1}{\path{#1}}}
\providecommand{\bibinfo}[2]{#2}
\ifx\xfnm\relax \def\xfnm[#1]{\unskip,\space#1}\fi
\bibitem[{Li et~al.(2018)Li, Simek, Lai, Daggett, Dagli, and Jones}]{li2018detection}
\bibinfo{author}{L.~Li}, \bibinfo{author}{O.~Simek}, \bibinfo{author}{A.~Lai}, \bibinfo{author}{M.~Daggett}, \bibinfo{author}{C.~K. Dagli}, \bibinfo{author}{C.~Jones},
\newblock \bibinfo{title}{Detection and characterization of human trafficking networks using unsupervised scalable text template matching},
\newblock in: \bibinfo{booktitle}{2018 IEEE International Conference on big data (big data)}, \bibinfo{organization}{IEEE}, \bibinfo{year}{2018}, pp. \bibinfo{pages}{3111--3120}.
\bibitem[{Lee et~al.(2021)Lee, Vajiac, Kulshrestha, Levy, Park, Jones, Rabbany, and Faloutsos}]{lee2021infoshield}
\bibinfo{author}{M.-C. Lee}, \bibinfo{author}{C.~Vajiac}, \bibinfo{author}{A.~Kulshrestha}, \bibinfo{author}{S.~Levy}, \bibinfo{author}{N.~Park}, \bibinfo{author}{C.~Jones}, \bibinfo{author}{R.~Rabbany}, \bibinfo{author}{C.~Faloutsos},
\newblock \bibinfo{title}{Infoshield: Generalizable information-theoretic human-trafficking detection},
\newblock in: \bibinfo{booktitle}{2021 IEEE 37th International Conference on Data Engineering (ICDE)}, \bibinfo{organization}{IEEE}, \bibinfo{year}{2021}, pp. \bibinfo{pages}{1116--1127}.
\bibitem[{Tong et~al.(2017)Tong, Zadeh, Jones, and Morency}]{tong-etal-2017-combating}
\bibinfo{author}{E.~Tong}, \bibinfo{author}{A.~Zadeh}, \bibinfo{author}{C.~Jones}, \bibinfo{author}{L.-P. Morency},
\newblock \bibinfo{title}{Combating human trafficking with multimodal deep models},
\newblock in: \bibinfo{editor}{R.~Barzilay}, \bibinfo{editor}{M.-Y. Kan} (Eds.), \bibinfo{booktitle}{Proceedings of the 55th Annual Meeting of the Association for Computational Linguistics (Volume 1: Long Papers)}, \bibinfo{publisher}{Association for Computational Linguistics}, \bibinfo{address}{Vancouver, Canada}, \bibinfo{year}{2017}, pp. \bibinfo{pages}{1547--1556}. \URLprefix \url{https://aclanthology.org/P17-1142/}. \DOIprefix\doi{10.18653/v1/P17-1142}.
\bibitem[{Vajiac et~al.(2023)Vajiac, Lee, Kulshrestha, Levy, Park, Olligschlaeger, Jones, Rabbany, and Faloutsos}]{vajiac2023deltashield}
\bibinfo{author}{C.~Vajiac}, \bibinfo{author}{M.-C. Lee}, \bibinfo{author}{A.~Kulshrestha}, \bibinfo{author}{S.~Levy}, \bibinfo{author}{N.~Park}, \bibinfo{author}{A.~Olligschlaeger}, \bibinfo{author}{C.~Jones}, \bibinfo{author}{R.~Rabbany}, \bibinfo{author}{C.~Faloutsos},
\newblock \bibinfo{title}{Deltashield: Information theory for human-trafficking detection},
\newblock \bibinfo{journal}{ACM Transactions on Knowledge Discovery from Data} \bibinfo{volume}{17} (\bibinfo{year}{2023}) \bibinfo{pages}{1--27}.
\bibitem[{Nair et~al.(2024)Nair, Liu, Vajiac, Olligschlaeger, Chau, Cazzolato, Jones, Faloutsos, and Rabbany}]{nair2024t}
\bibinfo{author}{P.~Nair}, \bibinfo{author}{J.~Liu}, \bibinfo{author}{C.~Vajiac}, \bibinfo{author}{A.~Olligschlaeger}, \bibinfo{author}{D.~H. Chau}, \bibinfo{author}{M.~Cazzolato}, \bibinfo{author}{C.~Jones}, \bibinfo{author}{C.~Faloutsos}, \bibinfo{author}{R.~Rabbany},
\newblock \bibinfo{title}{T-net: Weakly supervised graph learning for combatting human trafficking},
\newblock in: \bibinfo{booktitle}{Proceedings of the AAAI Conference on Artificial Intelligence}, volume~\bibinfo{volume}{38}, \bibinfo{year}{2024}, pp. \bibinfo{pages}{22276--22284}.
\bibitem[{Li et~al.(2022)Li, Nair, Pelrine, and Rabbany}]{li2022extracting}
\bibinfo{author}{Y.~Li}, \bibinfo{author}{P.~Nair}, \bibinfo{author}{K.~Pelrine}, \bibinfo{author}{R.~Rabbany},
\newblock \bibinfo{title}{Extracting person names from user generated text: Named-entity recognition for combating human trafficking},
\newblock in: \bibinfo{booktitle}{Findings of the Association for Computational Linguistics: ACL 2022}, \bibinfo{year}{2022}, pp. \bibinfo{pages}{2854--2868}.
\bibitem[{Perez et~al.(2023)Perez, Rivas, Turek, Sooksatra, Quevedo, Bichler, Cerny, Giddens, and Petter}]{perez2023decoding}
\bibinfo{author}{A.~R. Perez}, \bibinfo{author}{P.~Rivas}, \bibinfo{author}{J.~Turek}, \bibinfo{author}{K.~Sooksatra}, \bibinfo{author}{E.~Quevedo}, \bibinfo{author}{G.~Bichler}, \bibinfo{author}{T.~Cerny}, \bibinfo{author}{L.~Giddens}, \bibinfo{author}{S.~Petter},
\newblock \bibinfo{title}{Decoding the obfuscated: Advanced ner techniques on commercial sex advertisement data},
\newblock in: \bibinfo{booktitle}{2023 International Conference on Computational Science and Computational Intelligence (CSCI)}, \bibinfo{organization}{IEEE}, \bibinfo{year}{2023}, pp. \bibinfo{pages}{144--151}.
\bibitem[{Esfahani et~al.(2019)Esfahani, Cafarella, Pouyan, DeAngelo, Eneva, and Fano}]{esfahani2019context}
\bibinfo{author}{S.~S. Esfahani}, \bibinfo{author}{M.~J. Cafarella}, \bibinfo{author}{M.~B. Pouyan}, \bibinfo{author}{G.~DeAngelo}, \bibinfo{author}{E.~Eneva}, \bibinfo{author}{A.~E. Fano},
\newblock \bibinfo{title}{Context-specific language modeling for human trafficking detection from online advertisements},
\newblock in: \bibinfo{booktitle}{Proceedings of the 57th annual meeting of the association for computational linguistics}, \bibinfo{year}{2019}, pp. \bibinfo{pages}{1180--1184}.
\bibitem[{Zhu et~al.(2019)Zhu, Li, and Jones}]{zhu2019identification}
\bibinfo{author}{J.~Zhu}, \bibinfo{author}{L.~Li}, \bibinfo{author}{C.~Jones},
\newblock \bibinfo{title}{Identification and detection of human trafficking using language models},
\newblock in: \bibinfo{booktitle}{2019 European Intelligence and Security Informatics Conference (EISIC)}, \bibinfo{organization}{IEEE}, \bibinfo{year}{2019}, pp. \bibinfo{pages}{24--31}.
\bibitem[{Li et~al.(2023)Li, Tobey, Mayorga, Caltagirone, and {\"O}zalt{\i}n}]{li2023detecting}
\bibinfo{author}{R.~Li}, \bibinfo{author}{M.~Tobey}, \bibinfo{author}{M.~E. Mayorga}, \bibinfo{author}{S.~Caltagirone}, \bibinfo{author}{O.~Y. {\"O}zalt{\i}n},
\newblock \bibinfo{title}{Detecting human trafficking: Automated classification of online customer reviews of massage businesses},
\newblock \bibinfo{journal}{Manufacturing \& Service Operations Management} \bibinfo{volume}{25} (\bibinfo{year}{2023}) \bibinfo{pages}{1051--1065}.
\bibitem[{Saxena et~al.(2023)Saxena, Bashpole, Van~Dijck, and Spanakis}]{saxena2023idtraffickers}
\bibinfo{author}{V.~Saxena}, \bibinfo{author}{B.~Bashpole}, \bibinfo{author}{G.~Van~Dijck}, \bibinfo{author}{G.~Spanakis},
\newblock \bibinfo{title}{Idtraffickers: An authorship attribution dataset to link and connect potential human-trafficking operations on text escort advertisements},
\newblock \bibinfo{journal}{arXiv preprint arXiv:2310.05484}  (\bibinfo{year}{2023}).
\bibitem[{Giorgi et~al.(2021)Giorgi, Nitski, Wang, and Bader}]{giorgi2021declutrdeepcontrastivelearning}
\bibinfo{author}{J.~Giorgi}, \bibinfo{author}{O.~Nitski}, \bibinfo{author}{B.~Wang}, \bibinfo{author}{G.~Bader}, \bibinfo{title}{Declutr: Deep contrastive learning for unsupervised textual representations}, \bibinfo{year}{2021}. \URLprefix \url{https://arxiv.org/abs/2006.03659}. \href{http://arxiv.org/abs/2006.03659}{{\tt arXiv:2006.03659}}.
\bibitem[{Keskin et~al.(2021)Keskin, Bott, and Freeman}]{keskin2021cracking}
\bibinfo{author}{B.~B. Keskin}, \bibinfo{author}{G.~J. Bott}, \bibinfo{author}{N.~K. Freeman},
\newblock \bibinfo{title}{Cracking sex trafficking: Data analysis, pattern recognition, and path prediction},
\newblock \bibinfo{journal}{Production and Operations Management} \bibinfo{volume}{30} (\bibinfo{year}{2021}) \bibinfo{pages}{1110--1135}.
\bibitem[{Freeman et~al.(2025)Freeman, Bott, Keskin, and Marcantonio}]{Freeman-2025-multisite}
\bibinfo{author}{N.~K. Freeman}, \bibinfo{author}{G.~J. Bott}, \bibinfo{author}{B.~B. Keskin}, \bibinfo{author}{T.~L. Marcantonio},
\newblock \bibinfo{title}{A multi-site data sample for analyzing the online commercial sex ecosystem},
\newblock \bibinfo{journal}{Scientific Data} \bibinfo{volume}{12} (\bibinfo{year}{2025}). \DOIprefix\doi{https://doi.org/10.1038/s41597-025-04442-w}.
\bibitem[{Vaswani et~al.(2017)Vaswani, Shazeer, Parmar, Uszkoreit, Jones, Gomez, Kaiser, and Polosukhin}]{vaswani2017attention}
\bibinfo{author}{A.~Vaswani}, \bibinfo{author}{N.~Shazeer}, \bibinfo{author}{N.~Parmar}, \bibinfo{author}{J.~Uszkoreit}, \bibinfo{author}{L.~Jones}, \bibinfo{author}{A.~N. Gomez}, \bibinfo{author}{{\L}.~Kaiser}, \bibinfo{author}{I.~Polosukhin},
\newblock \bibinfo{title}{Attention is all you need},
\newblock \bibinfo{journal}{Advances in neural information processing systems} \bibinfo{volume}{30} (\bibinfo{year}{2017}).
\bibitem[{Devlin et~al.(2019)Devlin, Chang, Lee, and Toutanova}]{devlin-etal-2019-bert}
\bibinfo{author}{J.~Devlin}, \bibinfo{author}{M.-W. Chang}, \bibinfo{author}{K.~Lee}, \bibinfo{author}{K.~Toutanova},
\newblock \bibinfo{title}{{BERT}: Pre-training of deep bidirectional transformers for language understanding},
\newblock in: \bibinfo{booktitle}{Proceedings of the 2019 Conference of the North {A}merican Chapter of the Association for Computational Linguistics: Human Language Technologies, Volume 1 (Long and Short Papers)}, \bibinfo{publisher}{Association for Computational Linguistics}, \bibinfo{year}{2019}, pp. \bibinfo{pages}{4171--4186}. \DOIprefix\doi{10.18653/v1/N19-1423}.
\bibitem[{Liu et~al.(2019)Liu, Ott, Goyal, Du, Joshi, Chen, Levy, Lewis, Zettlemoyer, and Stoyanov}]{roberta-2019}
\bibinfo{author}{Y.~Liu}, \bibinfo{author}{M.~Ott}, \bibinfo{author}{N.~Goyal}, \bibinfo{author}{J.~Du}, \bibinfo{author}{M.~Joshi}, \bibinfo{author}{D.~Chen}, \bibinfo{author}{O.~Levy}, \bibinfo{author}{M.~Lewis}, \bibinfo{author}{L.~Zettlemoyer}, \bibinfo{author}{V.~Stoyanov}, \bibinfo{title}{{RoBERTa}: A robustly optimized {BERT} pretraining approach}, \bibinfo{year}{2019}. \URLprefix \url{https://arxiv.org/abs/1907.11692}. \href{http://arxiv.org/abs/1907.11692}{{\tt arXiv:1907.11692}}.
\bibitem[{Warner et~al.(2024)Warner, Chaffin, Clavié, Weller, Hallström, Taghadouini, Gallagher, Biswas, Ladhak, Aarsen, Cooper, Adams, Howard, and Poli}]{modernbert-2024}
\bibinfo{author}{B.~Warner}, \bibinfo{author}{A.~Chaffin}, \bibinfo{author}{B.~Clavié}, \bibinfo{author}{O.~Weller}, \bibinfo{author}{O.~Hallström}, \bibinfo{author}{S.~Taghadouini}, \bibinfo{author}{A.~Gallagher}, \bibinfo{author}{R.~Biswas}, \bibinfo{author}{F.~Ladhak}, \bibinfo{author}{T.~Aarsen}, \bibinfo{author}{N.~Cooper}, \bibinfo{author}{G.~Adams}, \bibinfo{author}{J.~Howard}, \bibinfo{author}{I.~Poli}, \bibinfo{title}{Smarter, better, faster, longer: A modern bidirectional encoder for fast, memory efficient, and long context finetuning and inference}, \bibinfo{year}{2024}. \URLprefix \url{https://arxiv.org/abs/2412.13663}. \href{http://arxiv.org/abs/2412.13663}{{\tt arXiv:2412.13663}}.
\bibitem[{Sp\"{a}rck~Jones(1972)}]{sparck1972statistical}
\bibinfo{author}{K.~Sp\"{a}rck~Jones},
\newblock \bibinfo{title}{A statistical interpretation of term specificity and its application in retrieval},
\newblock \bibinfo{journal}{Journal of Documentation} \bibinfo{volume}{28} (\bibinfo{year}{1972}) \bibinfo{pages}{11--21}.
\bibitem[{Pedregosa et~al.(2011)Pedregosa, Varoquaux, Gramfort, Michel, Thirion, Grisel, Blondel, Prettenhofer, Weiss, Dubourg, Vanderplas, Passos, Cournapeau, Brucher, Perrot, and Duchesnay}]{scikit-learn}
\bibinfo{author}{F.~Pedregosa}, \bibinfo{author}{G.~Varoquaux}, \bibinfo{author}{A.~Gramfort}, \bibinfo{author}{V.~Michel}, \bibinfo{author}{B.~Thirion}, \bibinfo{author}{O.~Grisel}, \bibinfo{author}{M.~Blondel}, \bibinfo{author}{P.~Prettenhofer}, \bibinfo{author}{R.~Weiss}, \bibinfo{author}{V.~Dubourg}, \bibinfo{author}{J.~Vanderplas}, \bibinfo{author}{A.~Passos}, \bibinfo{author}{D.~Cournapeau}, \bibinfo{author}{M.~Brucher}, \bibinfo{author}{M.~Perrot}, \bibinfo{author}{E.~Duchesnay},
\newblock \bibinfo{title}{Scikit-learn: Machine learning in {P}ython},
\newblock \bibinfo{journal}{Journal of Machine Learning Research} \bibinfo{volume}{12} (\bibinfo{year}{2011}) \bibinfo{pages}{2825--2830}.
\bibitem[{Reimers and Gurevych(2019)}]{reimers2019sentence}
\bibinfo{author}{N.~Reimers}, \bibinfo{author}{I.~Gurevych},
\newblock \bibinfo{title}{Sentence-bert: Sentence embeddings using siamese bert-networks},
\newblock \bibinfo{journal}{arXiv preprint arXiv:1908.10084}  (\bibinfo{year}{2019}).
\bibitem[{Kalamkar et~al.(2019)Kalamkar, Mudigere, Mellempudi, Das, Banerjee, Avancha, Vooturi, Jammalamadaka, Huang, Yuen et~al.}]{kalamkar2019study}
\bibinfo{author}{D.~Kalamkar}, \bibinfo{author}{D.~Mudigere}, \bibinfo{author}{N.~Mellempudi}, \bibinfo{author}{D.~Das}, \bibinfo{author}{K.~Banerjee}, \bibinfo{author}{S.~Avancha}, \bibinfo{author}{D.~T. Vooturi}, \bibinfo{author}{N.~Jammalamadaka}, \bibinfo{author}{J.~Huang}, \bibinfo{author}{H.~Yuen}, et~al.,
\newblock \bibinfo{title}{A study of bfloat16 for deep learning training},
\newblock \bibinfo{journal}{arXiv preprint arXiv:1905.12322}  (\bibinfo{year}{2019}).
\bibitem[{Wang et~al.(2024)Wang, Sun, Huang, and Rudin}]{wang2024dimensionreductionlocallyadjusted}
\bibinfo{author}{Y.~Wang}, \bibinfo{author}{Y.~Sun}, \bibinfo{author}{H.~Huang}, \bibinfo{author}{C.~Rudin}, \bibinfo{title}{Dimension reduction with locally adjusted graphs}, \bibinfo{year}{2024}. \URLprefix \url{https://arxiv.org/abs/2412.15426}. \href{http://arxiv.org/abs/2412.15426}{{\tt arXiv:2412.15426}}.
\bibitem[{McInnes et~al.(2017)McInnes, Healy, and Astels}]{McInnes2017}
\bibinfo{author}{L.~McInnes}, \bibinfo{author}{J.~Healy}, \bibinfo{author}{S.~Astels},
\newblock \bibinfo{title}{hdbscan: Hierarchical density based clustering},
\newblock \bibinfo{journal}{The Journal of Open Source Software} \bibinfo{volume}{2} (\bibinfo{year}{2017}) \bibinfo{pages}{205}. \DOIprefix\doi{10.21105/joss.00205}.
\bibitem[{Administration(2021)}]{dea2021emoji}
\bibinfo{author}{D.~E. Administration}, \bibinfo{title}{Emoji drug code: Decoded}, \bibinfo{year}{2021}. \URLprefix \url{https://www.dea.gov/sites/default/files/2021-12/Emoji%20Decoded.pdf}, \bibinfo{note}{accessed: 2025-03-26}.
\bibitem[{DarkOwl(2022)}]{darkowl2022emoji}
\bibinfo{author}{DarkOwl}, \bibinfo{title}{The dark side of emojis: Exploring emoji use in illicit and underground activities}, \bibinfo{year}{2022}. \URLprefix \url{https://www.darkowl.com/blog-content/the-dark-side-of-emojis-exploring-emoji-use-in-illicit-and-underground-activities/}, \bibinfo{note}{accessed: 2025-03-26}.

\end{thebibliography}

\end{document}